\documentclass[letterpaper, 10 pt, conference]{ieeeconf}  

\usepackage{cite}
\usepackage{amsmath,amssymb,amsfonts}
\usepackage{algorithmic}
\usepackage{graphicx}
\usepackage{textcomp}
\usepackage{xcolor}
\usepackage{caption}
\usepackage{subcaption}
\usepackage{bm}
\usepackage[english]{babel}
\usepackage{cancel}
\usepackage{wrapfig}
\usepackage{verbatim} 
\usepackage{fancyhdr}

\makeatletter
\def\maketag@@@#1{\hbox{\m@th\normalfont\normalsize#1}}
\makeatother

\usepackage[bookmarks=true, colorlinks=true]{hyperref}

\usepackage[linesnumbered,ruled,vlined]{algorithm2e}

\newtheorem{theorem}{Theorem}

\newtheorem{corollary}{Corollary}

\newcommand{\prob}{\ensuremath{\mathbb{P}}}
\newcommand{\Exp}{\ensuremath{\mathbb{E}}}
\newcommand{\br}{b'^-} 
\newcommand{\SPB}[1]{b^{-}_{#1}} 
\newcommand{\SB}[1]{b_{#1}} 
\newcommand{\TM}{\prob_T(s_{k+1}|s_k,a_k)}
\newcommand{\OM}{\prob_O(o_{k+1}|s_{k+1})} 
\newcommand{\MTT}[1]{} 

\newcommand{\TRE}[1]{\tau_{suffix}^{#1}}

\IEEEoverridecommandlockouts                              

\title{\LARGE \bf
Previous Knowledge Utilization In Online Anytime Belief Space Planning
}

\author{Michael Novitsky, Moran Barenboim and Vadim Indelman 
\thanks{Michael Novitsky and Moran Barenboim are with the Technion Autonomous Systems Program (TASP), Technion - Israel Institute of Technology, Haifa 32000,
	Israel, {\tt miken1990@campus.technion.ac.il; moranbar@campus.technion.ac.il}. Vadim Indelman is with the Department of Aerospace Engineering and with the Department of Data and Decision Science, Technion - Israel Institute of Technology, Haifa 32000, Israel. {\tt vadim.indelman@technion.ac.il}. This work was  partially supported by US NSF/US-Israel BSF.}%
}

\begin{document}

\maketitle

\thispagestyle{fancy}
\fancyhf{} 
\fancyfoot[L]{\footnotesize This work has been submitted to the IEEE for possible publication. Copyright may be transferred without notice, after which this version may no longer be accessible.} 


\begin{abstract}
Online planning under uncertainty remains a critical challenge in robotics and autonomous systems. While tree search techniques are commonly employed to construct partial future trajectories within computational constraints, most existing methods discard information from previous planning sessions considering  continuous spaces. This study presents a novel, computationally efficient approach that leverages historical planning data in current decision-making processes. We provide theoretical foundations for our information reuse strategy and introduce an algorithm based on Monte Carlo Tree Search (MCTS) that implements this approach. Experimental results demonstrate that our method significantly reduces computation time while maintaining high performance levels. Our findings suggest that integrating historical planning information can substantially improve the efficiency of online decision-making in uncertain environments, paving the way for more responsive and adaptive autonomous systems.
\end{abstract}

\section{INTRODUCTION}
Autonomous agents often operate under uncertainty due to sensor noise and incomplete information, maintaining a belief (probability distribution) over possible states instead of direct access to the true environment state. Partially Observable Markov Decision Processes (POMDPs) provide a framework for such settings, but solving them optimally is computationally intractable (PSPACE-complete) \cite{Papadimitriou87math}, mainly due to the curse of history, curse of dimensionality, and continuous state, action and observation spaces common in real-world applications.

Recent advancements have introduced online algorithms \cite{Silver10nips} \cite{Somani13nips} \cite{Hoerger21icra} that find approximate solutions to POMDPs.
These algorithms operate within limited budget constraints, such as restricted time, and employ a sampling-based approach to construct partial trees and search for the optimal action that maximizes the expected cumulative reward. By sampling a subset of the belief space, these algorithms effectively address both the curse of history and the curse of dimensionality, which are key obstacles in solving POMDPs.
 
In POMDPs, the reward function of a belief node is typically formulated as the expected reward over states. However, this formulation may be insufficient for certain problems, such as information gathering and active sensing. In such cases, the problem is commonly addressed as Belief Space Planning (BSP) or $\rho$-POMDP \cite{AryaLopez10nips}, where the reward is defined over the belief itself.
Information-theoretic measures, such as information gain and differential entropy, are commonly used to quantify uncertainty in the decision-making process \cite{Fischer20icml}. However, exact calculation of information-theoretic rewards becomes intractable for general distributions, as it requires integrating over all possible states. To address this challenge, approximation methods such as kernel density estimation (KDE) and particle filter estimation \cite{Boers10fusion} have been proposed in the literature. Nonetheless, these methods still incur significant computational expenses, with computation complexity scaling quadratically with the number of samples. As reward calculation is performed for each node in the tree, it becomes the primary source of computational complexity in online planning algorithms.

The main objective of this paper is to improve planning efficiency within a non-parametric setting, continuous state, action and observation spaces, and general reward functions. To address these challenges, we contribute a novel approach that leverages the Multiple Importance Sampling framework \cite{Veach95siggraph} to tackle the problem of reusing information from previous planning sessions.
Our approach introduces a new algorithm specifically designed to utilize knowledge gathered during prior planning sessions. We demonstrate how our method can be integrated with Monte Carlo Tree Search (MCTS) to create a novel online algorithm called Incremental Reuse Particle Filter Tree (IR-PFT).
We evaluate our algorithm in an online planning setting, demonstrating reduced planning time without performance loss.

The code for this paper is available at \href{https://github.com/miken1990/ir-pft}{https://github.com/miken1990/ir-pft}.

\section{RELATED WORK}
Solving POMDPs is challenging, but recent advances, such as the POMCP algorithm \cite{Silver10nips}, have made significant progress. POMCP extends the UCT algorithm \cite{Kocsis06ecml} to handle partial observability. During each simulation, a state particle is sampled from the current belief, propagated through the search tree, and information like visitation count and accumulated reward is recorded. Action selection follows a Multi-Armed Bandit approach. POMCP assumes discrete state, action and observation spaces, and a state-based reward function. The number of samples at each belief node depends on the number of simulations it has participated in, with less visited nodes having fewer samples.
POMCPOW \cite{Sunberg18icaps} extends POMCP to continuous action and observation spaces by using progressive widening and representing beliefs as weighted particle sets. It assumes access to an observation likelihood model, where each simulated state is added to the weighted belief, and a new state is sampled based on its weight.
In PFT-DPW \cite{Sunberg18icaps} the authors adopt particle filter formulation for belief update and each belief is represented with a constant number of samples. 
\cite{Thomas21arxiv} introduce $\rho-$POMCP which propagates a set particles in each simulation using particle filter and adds it to existing particles in visited nodes. 
Frequently visited nodes achieve better representation, with convergence proven asymptotically for continuous, bounded $\rho$.
\cite{Fischer20icml} introduce the IPFT algorithm, which extends PFT \cite{Sunberg18icaps}. They use a reward defined as a linear combination of differential entropy and expected state reward. In each simulation, a particle set is sampled from the root belief and propagated through the tree. Entropy estimates are averaged across particle sets at each belief node to estimate differential entropy.
\cite{Hoerger21icra} propose LABECOP for continuous observation spaces. At each belief node $b$, a state particle $s$ is sampled, an action $a$ is chosen using modified UCB, and an observation $o$ is sampled. Previous states from $b,a$ are reweighted by $o$ to improve value function estimate $\hat{Q}(b,a)$. 
SITH-BSP\cite{Sztyglic22iros, Zhitnikov24ijrr}  and AI-FSSS\cite{Barenboim22ijcai} make use of simplification of reward function calculation and observation space sampling accordingly, while preserving action consistency. 
\cite{Zhitnikov22ai} quantify the effect of applying simplification and extend $\rho$-POMDP to $\mathbb{P}\rho-$POMDP, while providing stochastic bounds on the return. 
DESPOT \cite{Somani13nips} and subsequent works \cite{Ye17jair}, \cite{Garg19rss}, \cite{Cai21ijrr} propose algorithms that use determinized random sampling to build the search tree incrementally, with recent work addressing large observation spaces \cite{Garg19rss}. The use of $\alpha$ - vectors in \cite{Ye17jair}, \cite{Garg19rss}, \cite{Cai21ijrr} restricts their application to POMDPs with state-dependent reward functions.
Previous methods start each planning session from scratch, while iX-BSP \cite{Farhi19icra}, \cite{Farhi21arxiv} proposes reuse but assumes an open loop setting and doesn't address non-parametric beliefs. In this work, we address continuous state, action, and observation spaces with general belief-dependent rewards, a non-parametric framework, and a closed-loop setting.

\section{NOTATIONS} \label{notations section}

\subsection{POMDP}
POMDP is a 7-tuple  $(S, A, O, \prob_T, \prob_O, r, b_0)$,  where $S$, $A$ and $O$ correspond to state, action and observation spaces. $\TM$ is the state transition density function, $\OM$ is the observation density function, $r(b,a,b')$ represents the reward function based on the current belief $b$, the action $a$, and the subsequent belief $b'$, while $b_0$ denotes the current belief over states. 
We denote by $H_{k}=(b_0,a_0,o_1,..,o_k)=\{b_0,o_{1:k},a_{1:k-1}\}$ the history up to time $k$, which consists of a series of actions made and observations received. 
Since the exact state of the world is not known and we only receive observations, a probability distribution (belief) over states is maintained $b_k=\prob(s_k|H_{k})$. It is assumed that the belief is sufficient statistics for the decision making and a Bayesian update is used to update the belief recursively: 
\begin{align} \label{eq:belief_update}
	b_{k+1}=\eta \OM \int_{s_k} \TM b_k ds_{k}.
\end{align} 
where $\eta$ is a normalization term. 
A policy $\pi\in \Pi$ is a mapping from belief space to action space $\pi : b \rightarrow a$.  We define the value function $V^\pi$  for any policy $\pi$ and horizon $d$ as
\begin{align} \label{eq:policy_value_function}
V^{\pi}(b_k)=\underset{b_{k+1:k+d}}{\Exp}[G_k|b_k,\pi].
\end{align} 
where $\pi \triangleq \pi_{k:k+d-1}$ represents a sequence of policies for horizon d and $G_k=\sum^{k+d-1}_{i=k}r(b_i,\pi_{i}(b_i), b_{i+1})$ 
is the return. Similarly,
we define the action value function $Q^{\pi}$ as
\begin{align} \label{eq:action_value_function}
	Q^{\pi}(b_k,a)=\Exp_{b_{k+1}}[r(b_k,a,b_{k+1})+V^{\pi}(b_{k+1})].
\end{align} 
\subsection{Non-Parametric Setting}\label{sec:Nonparambeliefs}
In our work we assume a non-parametric setting, where we use collections of state particles to estimate complex belief distributions. 
We leverage the particle filter method \cite{Thrun05book} 
to update our approximations of posterior distributions as we receive new observations from the environment. 
The theoretical belief $b_k$ is approximated using $m$ particles $\{s^i_k\}^m_{i=1}$, assuming resampling at each particle filter step, which ensures uniform weights of $\frac{1}{m}$
\begin{align} \label{eq:belief_approximation_resampled}
	\hat{b}_k = \frac{1}{m} \sum^{m}_{i=1}{\delta(s-s^i_k)}. \hfill
\end{align} 
Given resampled belief $\hat{b}_k$, action $a_k$, and propagated belief $\hat{b}^-_{k+1}$, calculating $\prob(\hat{b}^-_{k+1}|\hat{b}_k, a_k)$ involves determining all the matchings between the states in $\hat{b}_k$ and those in $\hat{b}^-_{k+1}$ which is $\sharp P$-complete \cite{Valiant1979}.
We assume, similar to \cite{Lim23jair}, that the beliefs are not permutation invariant, meaning particle beliefs with different particle orders are not considered identical. This assumption simplifies the derivation of the propagated belief likelihood. Consequently, we can express $\hat{b}_k$ as $\{s^{i}_{k},\frac{1}{m}\}^m_{i=1}$ and $\hat{b}^-_{k+1}$ as $\{s^{-i}_{k+1},\frac{1}{m}\}^m_{i=1}$
\begin{align} \label{propagated_belief_likelihood}
	\prob(\hat{b}^-_{k+1}|\hat{b}_k, a_k)= \frac{1}{m} \prod_{i=1}^{m} 	\prob(s^{-i}_{k+1}|s^{i}_{k}, a_k).
\end{align}
In the rest of the paper we assume a non-parametric setting and for the ease of notation we remove the hat sign $\hat{}$ from all beliefs. 

\subsection{Importance Sampling} \label{subsection:importance_sampling}
Importance sampling estimates properties of a target distribution $p(x)$ by sampling from a proposal distribution $q(x)$, assigning weights to adjust each sample's contribution according to $p(x)$
\begin{align} \label{eq:importance_sampling}
	\hat{\Exp}^{IS}_p[f(x)] =
	\frac{1}{N}\sum^{N}_{i=1}w^i \cdot f(x^i), \ w^i=\frac{p(x^i)}{q(x^i)}, \ x^i \sim q.
\end{align}
The distribution $q$ must satisfy $q(x^i)=0 \Rightarrow p(x^i)=0$.
With $M$ proposal distributions $\{q_m\}^M_{m=1}$, Multiple Importance Sampling formulation \cite{Veach95siggraph} can be used:
\begin{align} \label{eq:multiple_importance_sampling}
	\hat{\Exp}^{MIS}_p[f(x)] =  \sum^{M}_{m=1}\frac{1}{n_m}\sum^{n_m}_{i=1}w^{m}(x^{i,m})\frac{p(x^{i,m})}{q_{m}(x^{i,m})}f(x^{i,m}).
\end{align}
Here, $n_m$ denotes the number of samples that originate from distribution $q_m$, $x^{i,m}$ denotes the $i$th sample that originates from distribution $q_m$ and the weights $w^{m}$ must satisfy 
\begin{align} \label{eq:weights_mis}
	q_{m}(x^{i,m}) = & 0 \Rightarrow w^m(x)f(x)p(x) = 0.  \\
	f(x^{i,m}) \neq & 0 \Rightarrow \sum^M_{m=1}w^m(x^{i,m})=1. \notag
\end{align}
We assume that the weights $w^m$ are determined using the balance heuristic which bounds the variance of the estimator \cite{Veach95siggraph} and in this case the MIS estimator is
\begin{align} \label{mis_balance_h}
	\hat{\Exp}^{MIS}_p[f(x)]=\sum^{M}_{m=1}\sum^{n_m}_{i=1}\frac{p(x^{i,m})}{\sum^M_{j=1} n_j \cdot q_{j}(x^{i,m})} f(x^{i,m}).
\end{align}

\subsection{PFT-DPW}
The PFT-DPW algorithm \cite{Sunberg18icaps} is based on the UCT algorithm \cite{Kocsis06ecml}  and expands its application to a continuous state, action and observation setting. It utilizes Monte-Carlo simulations to progressively construct a policy tree for the belief MDP \cite{Sunberg18icaps}. 
At every belief node $b_k$ and action $a_k$ it sets up visitation counts $N(b_k,a_k)$ and $N(b_k)$, where $N(b_k)=\sum_{a_k}N(b_k,a_k)$ and action-value function is calculated incrementally 
\begin{align} \label{q_mcts}
	Q(b_k,a_k) \triangleq \frac{1}{N} \sum^{N}_{i=1}{G^i_k},
\end{align}
by averaging accumulated reward upon initiating from node $b_k$ and taking action $a_k$ within the tree. Notably, $Q(b_k,a_k)$ \eqref{q_mcts} is not equal to $Q^{\pi}(b_k,a_k)$ \eqref{eq:action_value_function} 
as the policy varies across different simulations within the tree, causing the distribution of the trajectories to be non-stationary, hence the absence of the $\pi$ superscript.
The particle filter generates a propagate belief $b^-_{k+1}$ and posterior belief $b_{k+1}$ from $b_k$ and $a_k$, sampling observation $o_{k+1}$ and computing reward r
\begin{align} \label{particle_filter_update}
	\SB{k+1}, \SPB{k+1}, o_k, r \leftarrow G_{PF(m)}(b_k,a_k).
\end{align}
To handle continuous spaces, Double Progressive Widening limits a node's children to $kN^{\alpha}$, where $N$ is the node visit count, and $k$ and $\alpha$ are hyperparameters \cite{Sunberg18icaps}.

\section{APPROACH}
Our contributions are threefold: (1) an efficient incremental update method for the Multiple Importance Sampling (MIS) estimator, enabling action-value estimation from prior and newly arriving data; (2) the application of MIS for experience-based value estimation using expert-provided data without planning; and (3) an MCTS-inspired online algorithm that speeds up computations by reusing data from previous planning sessions.
\subsection{Incremental Multiple Importance Sampling Update}
In our setting, samples arrive incrementally in batches. A straightforward computation of \eqref{mis_balance_h} would necessitate a complexity of $O(M^2 \cdot n_{avg})$, where $M$ denotes the number of different distributions and $n_{avg}$ denotes the average sample count across all distributions. We develop an efficient way to update the estimator \eqref{mis_balance_h} incrementally in the theorem below.
\begin{theorem} \label{th:incremental_mis_update}
	\textit{Consider an MIS estimator \eqref{mis_balance_h} with $M$ different distributions and $n_m$ samples for each distribution $q_m \in \{q_1,..., q_M\}$. Given a  batch of $L$ I.I.D samples from distribution $q_{m'}$, where $q_{m'}$ could be one of the existing distributions or a new, previously unseen distribution, $\hat{\Exp}_p^{MIS}[f(x)]$ \eqref{mis_balance_h} can be efficiently updated with a computational complexity of O($M\cdot n_{avg} + M \cdot L$) and memory complexity O($M \cdot n_{avg}$).} 
	\small
\end{theorem}
\textit{Proof. see Appendix \ref{th1_proof}.}
\subsection{Experience-Based Value Function Estimation}\label{subs:experience_based_value_function_estimation}
We assume that we have access to a dataset 
\begin{align} \label{dataset}
	D \triangleq \{\tau^i, G^i_{k_i}\}^{|D|}_{i=1}
\end{align}
of trajectories executed by an agent that followed a policy $\pi$. Each trajectory is defined as the sequence 
\begin{align} \label{trajectory_def}
	\tau^{i} \triangleq  (b^{i}_{k_i}, a^i_{k_i}) & \rightarrow (b^{-i}_{k_i+1}, o^i_{k_i+1}, b^{i}_{k_i+1}a^i_{k_i+1}) \rightarrow ... \nonumber \\ & \rightarrow (b^{-i}_{k_i+d},o^{i}_{k_i+d}, b^{i}_{k_i+d}), \hfill
\end{align}		
where $k_i$ represents the starting time index and is used to differentiate between different steps in trajectory $\tau^i$ and $d$ is the horizon length. We assume that the agent applied a particle filter with resampling at each step of the trajectory.
The return $G^i$  associated with trajectory $\tau^i$ is defined as the accumulated reward,
\begin{align} \label{trajectory_return_def}
	G^{i}_{k_i} \triangleq \sum_{j=0}^{d-1}r(b^i_{k_i+j},a^i_{k_i+j}, b^i_{k_i+j+1}).
\end{align}
In this section, we evaluate $V^\pi(b_k)$ for the current belief $b_k$ using only the dataset $D$ \eqref{dataset}, without planning. Such estimation is important in data-expensive domains like autonomous vehicles \cite{nuplan} and robotic manipulation tasks \cite{m2019scaling}.
In the next section, we will expand our methodology to include planning.

Reusing trajectories where the initial belief is set to $b_k$ presents no challenge - we can aggregate all trajectories that begin with belief $b_k$ and action $a_k$ and assuming we have $N$ such trajectories, we define a sample-based estimator 
\begin{align}\label{simple_reuse}
	\hat{Q}^{\pi}(b_k,a_k) \triangleq \frac{1}{N} \sum^{N}_{i=1}{G^i_{k}}.
\end{align}
However, in continuous state, action and observation spaces, the probability of sampling the same belief twice is zero. Consequently, each trajectory in the dataset $D$ \eqref{dataset} will have an initial belief that is different from $b_k$.

\begin{figure} [!ht]
	\begin{center}
		\centering
		\includegraphics[width=0.35\columnwidth]{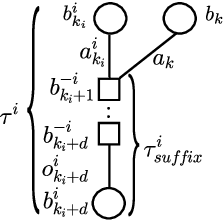}
		\scriptsize
		\caption{\scriptsize $\tau^i$ is a trajectory that was executed by an agent that followed policy $\pi$, $\tau^i_{suffix}$ is the part that we reuse from $\tau^i$ for the current belief $b_k$ and action $a_k$.}
		\normalsize
		\label{fig:trajectory_reuse}
	\end{center}
\end{figure}

To be able to reuse trajectories from \eqref{dataset}, we discard the initial belief and action of the trajectory, instead linking the current belief and action to the remainder of the trajectory.
Formally, given a trajectory $\tau^{i} \in D$, 
$\tau^{i} = (b^i_{k_i},a^i_{k_i})\rightarrow \TRE{i}$ where 
\begin{align} \label{eq:partial_traj}
	\TRE{i} \triangleq & (b^{-i}_{k_i+1},o^{i}_{k_i+1} ,b^{i}_{k_i+1},a^i_{k_i+1}) \rightarrow ... \nonumber \\ & \rightarrow (b^{-i}_{k_i+d},o^{i}_{k_i+d},b^{i}_{k_i+d}). \hfill 
\end{align}
and the current belief $b_k$ and action  $a_k$, we construct a new trajectory $\tau'_{i}$ (see Figure \ref{fig:trajectory_reuse}),
\begin{align} \label{eq:reuse_traj}
	\tau'_{i} \triangleq  (b_k, a_k)\rightarrow \TRE{i}.
\end{align}
To estimate $Q^{\pi}(b_k,a_k)$ using the information within trajectory $\tau^i$, two adjustments are required. Firstly, we need to modify the initial term in the return $G^i$ to be equal to $r(b_k, a_{k}, b^i_{k_i+1})$, recognizing that $b_k \neq b^i_{k_i}$ and $a_k \neq a^i_{k_i}$. Consequently, we define the return of trajectory $\tau'_{i}$ 
\begin{align} \label{eq:correct_g_reuse}
	\tilde{G}^i_{k} \triangleq G^i_{k_i} - r(b^i_{k_i}, a^i_{k_i}, b^i_{k_i+1}) + r(b_k, a_k, b^i_{k_i+1}).
\end{align}
Secondly, we need to adjust the weight of $\tilde{G}^i$ due to the disparity between $\prob(\TRE{i}|b^{i}_{k_i},a^i_{k_i}, \pi)$ and $\prob(\TRE{i}|b_k, a_k, \pi)$, which is acheived through importance sampling. The distribution $\prob(\cdot |b^{i}_{k_i},a^i_{k_i}, \pi)$ of partial trajectory $\TRE{i}$ is determined by the initial belief $b^{i}_{k_i}$ and action $a^i_{k_i}$. Given $N_{IS}$ partial trajectories sampled from the same distribution $\prob(\cdot |b^{i}_{k_i},a^i_{k_i}, \pi)$, we define an  Importance Sampling estimator 
\begin{align}\label{eq:is_reuse}
	\hat{Q}_{IS}^{\pi}(b_k,a_k) \triangleq \frac{1}{N_{IS}} \sum^{N_{IS}}_{i=1}{w^i \cdot \tilde{G}^i_{k}}.
\end{align}
where $w^i \triangleq \frac{\prob(\TRE{i}|b_k, a_k, \pi)}{\prob(\TRE{i}|b^{i}_{k_i},a^i_{k_i}, \pi)}$. 

As a result of our approach to constructing reusable trajectories as described in \eqref{eq:reuse_traj}, we can efficiently calculate the weights $w^i$ utilizing the theorem presented below.
\begin{theorem} \label{th:trajectory_likelihood_rat}
	\textit{Given belief node $b_k$, action $a_k$ and trajectory $\tau^i=(b^i_{k_i}, a^i_{k_i}) \rightarrow  \TRE{i}$ where $\TRE{i}$ is defined in \eqref{eq:partial_traj}, the following equality holds: 
	\begin{align}\label{eq:lklbelratios}
		\frac{\prob(\TRE{i}|b_k, a_k, \pi)}{\prob(\TRE{i}|b^i_{k_i},a^i_{k_i}, \pi)} = \frac{\prob(b^{-i}_{k_i+1}|\SB{k}, a_k)}{\prob(b^{-i}_{k_i+1}|b^i_{k_i}, a^i_{k_i})}.
	\end{align}	
	}
\end{theorem}
\textit{Proof. see appendix \ref{th:trajectory_likelihood_rat_proof}.}

We denote by $M$ the number of unique distributions of partial trajectories $\{\prob(\cdot|b^m_{k_m},a^m_{k_m}, \pi)\}^M_{m=1}$, where each distribution is defined by the initial belief $b^m_{k_m}$ and action $a^m_{k_m}$.
Additionally, we denote the sample count from each distribution as $n_m$.  Consequently, we can reformulate the dataset $D$ \eqref{dataset} as follows:
\begin{align} \label{dataset_two_idx}
	D \triangleq \{\tau^{l,m}, G^{l,m}_{k}\}^{M, n_m}_{m=1, l=1}.
\end{align}
Using this formulation, 
we define a multiple importance sampling estimator assuming the balance heuristic \eqref{mis_balance_h}, 
\begin{align} \label{mis_balance_h_suffix}
	\hat{Q}^{\pi}_{MIS}(b_k,a_k) \triangleq \sum^{M}_{m=1}\sum^{n_m}_{l=1}\frac{\prob(\TRE{l,m}|b_k, a_k, \pi) \tilde{G}^{l,m}_{k}}{\sum^M_{j=1} n_j \cdot \prob(\TRE{l,m}|b^j_{k_j},a^j_{k_j}, \pi)}.
\end{align}
where $\TRE{l,m}$ represents the $l$th partial trajectory that was sampled from the distribution $\prob(\cdot | b^{m}_{k_m}, a^{m}_{k_m},\pi)$
and $\tilde{G}^{l,m}_k$ is the adjusted accumulated reward \eqref{eq:correct_g_reuse}.

\begin{figure} [!ht]
	\begin{center}
		\centering
		\includegraphics[width=0.6\columnwidth]{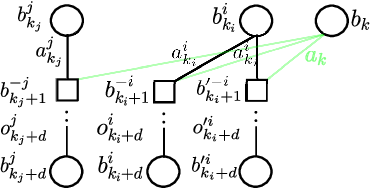}
		\scriptsize
		\caption{\scriptsize Illustration of reuse of three trajectories.}
		\normalsize
		\label{fig:forest_reuse}
	\end{center}
\end{figure}

Using Theorem \ref{th:trajectory_likelihood_rat}, we can re-write the MIS estimator \eqref{mis_balance_h_suffix} 
\begin{align} \label{eq:mis_est_prop_belief}
	\hat{Q}^{\pi}_{MIS}(\SB{k},a_k) \triangleq \sum^{M}_{m=1} 
	\sum^{n_m}_{l=1} \frac{\prob(b^{-l,m}_{k_{m}+1}|b_k,a_k)}{\sum^{M}_{j=1} n_j \cdot  \prob(b^{-l,m}_{k_{m}+1}|b^{j}_{k_{j}},a^{j}_{k_{j}})} \cdot \tilde{G}^{l,m}_k.
\end{align}
Since each element in the second sum of \eqref{eq:mis_est_prop_belief} corresponds to a propagated belief, which might appear more than once, we can rewrite the sum in a more compact form. Specifically, we group the terms based on unique propagated beliefs and account for their multiplicity:
\begin{align} \label{eq:mis_est_prop_belief_same}
	\hat{Q}^{\pi}_{MIS}(\SB{k},a_k) \triangleq \sum^{M}_{m=1} 
	\sum^{|C(b_{k_m},a_{k_m})|}_{l=1}W(b^{-l,m}_{k_{m}+1}) \cdot \sum_{y=1}^{N(b^{-l,m}_{k_{m}+1})} \tilde{G}^{m,l,y}_k.
\end{align}
The weights $W(b^{-l,m}_{k_{m}+1})$ are defined by:
\begin{align} \label{eq:mis_est_prop_belief_same_w}
	 W(b^{-l,m}_{k_{m}+1})=\frac{ \prob(b^{-l,m}_{k_{m}+1}|b_k,a_k)}{\sum^{M}_{j=1} n_j \cdot  \prob(b^{-l,m}_{k_{m}+1}|b^{j}_{k_{j}},a^{j}_{k_{j}})}
\end{align}
$C(b_{k_m},a_{k_m})$ denotes the set of reused propagated belief children associated with $b_{k_m}$ and $a_{k_m}$.
The term $N(b^{-l,m}_{k_{m}+1})$ represents the visitation count of $b^{-l,m}_{k_{m}+1}$, indicating the number of trajectories that pass through the propagated belief $b^{-l,m}_{k_{m}+1}$ and $\tilde{G}^{m,l,y}_k$ is the return of the $y$-th trajectory passing through $b^{-l,m}_{k_{m}+1}$.
Note that $b^{-l,m}_{k_{m}+1}$ in \eqref{eq:mis_est_prop_belief_same} represents unique propagated beliefs, which differs from \eqref{eq:mis_est_prop_belief}, where it denotes the propagated belief associated with a single trajectory.
Figure \ref{fig:forest_reuse} illustrates the estimator from \eqref{eq:mis_est_prop_belief_same}. For the current belief $b_k$ and action $a_k$, three prior trajectories are incorporated: two from ($b^i_{k_i}, a^{i}_{k_i}$) and one from ($b^j_{k_j}, a^{j}_{k_j}$). Light green edges show the connections between ($b_k,a_k$) and the reused nodes for estimating $\hat{Q}^{\pi}_{MIS}(\SB{k},a_k)$.
Further in this work, we consider a framework where the dataset $D$ \eqref{dataset_two_idx} expands over time with trajectory samples from an agent following policy $\pi$.  Theorem \ref{th:incremental_mis_update} is used to efficiently update the estimator \eqref{eq:mis_est_prop_belief} with new samples.

To clarify, our framework differs from standard off-policy evaluation methods. Traditional importance sampling for experience-based value estimation operates within the off-policy paradigm \cite{Sutton18book}, where trajectories originate from the current belief $b_k$, using behavioral ($\pi_b$) and target ($\pi_t$) policies to estimate $V^{\pi_t}(b_k)$. In contrast, we estimate $V^{\pi}(b_k)$ for the current belief and a specified policy $\pi$, with trajectories drawn form \emph{different} beliefs in the  dataset $D$ \eqref{dataset}.
To our knowledge, such a setting has not been addressed before in the context of action-value function estimation in POMDPs. 
\subsection{Our POMDP Planning Algorithm: IR-PFT}
Up to this point, we considered a specific single  policy, denoted as $\pi$, and utilized previously-generated  trajectories by an agent following $\pi$ to estimate the action-value function $Q^{\pi}(b_k,a_k)$. 
In this section, we present an anytime POMDP planning algorithm that uses trajectories from the dataset $D$, which includes data from previous planning sessions, to accelerate current planning.

We name our algorithm Incremental Reuse Particle Filter Tree (IR-PFT).
Instead of calculating $Q(b_k,a_k)$ from scratch in each planning session, we use previous experience to speed up the calculations.

We adopt the same approach as in Section \ref{subs:experience_based_value_function_estimation} to reuse trajectories, with three key modifications: first, the propagated belief nodes from the previous planning session in dataset $D$ have a shorter planning horizon. We extend the horizon of these nodes before reusing them;
second, the policy varies across different simulations (as in standard MCTS), resulting in a non-stationary distribution of reused trajectories in $D$; and third, we integrate the planning and generation of new trajectories with the reuse of previous trajectories within an anytime MCTS setting. 

\vspace{-3pt}
\setlength{\belowcaptionskip}{-1pt}
\begin{figure} [!ht]
	\begin{center}
		\centering
		\includegraphics[width=0.38\columnwidth]{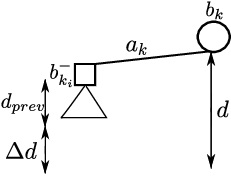}
		\scriptsize
		\caption{\scriptsize
		Illustration of horizon gap.}
		\normalsize
		\label{fig:horizon_gap_illustration}
	\end{center}
\end{figure}
Figure \ref{fig:horizon_gap_illustration} visually illustrates the horizon alignment process, where a propagated belief node $b^-_{k_i}$ with horizon $d_{prev}$ must be extended by $\Delta d$ to match the current horizon $d$.
We analyze the complexity of the correction of belief nodes from previous planning sessions in case of using the MCTS \cite{Auger13ecml} algorithm in Corollary \ref{cor:mcts_correction}.
\begin{corollary} \label{cor:mcts_correction}
	\textit{Given an MCTS tree  $T$ with horizon $d_{prev}$, number of simulations $m$ and $N$ nodes, extending its horizon by $\Delta d$ will require adding at most $m\cdot \Delta d$ nodes and reward calculations}.
	\small
\end{corollary}
The proof is straightforward: after $m$ simulations, the MCTS tree contains at most $m$ leaves and we need to extend each leaf by $\Delta d$ and for each new node we calculate a reward. 

We now define the estimator
\begin{align} \label{eq:mis_est_prop_belief_alg}
	\hat{Q}_{MIS}(\SB{k},a_k) \triangleq \sum^{M}_{m=1} 
	\sum^{|C(b_{k_m},a_{k_m})|}_{l=1}W(b^{-l,m}_{k_{m}+1}) \cdot \sum_{s=1}^{N(b^{-l,m}_{k_{m}+1})} \bar{G}^{m,l,y}_k,
\end{align}
where the weights $W(b^{-l,m}_{k_{m}+1})$ are defined in
\eqref{eq:mis_est_prop_belief_same_w}. $\bar{G}^{m,l,y}_k$ is the extended return defined as
\begin{align} \label{eq:extended_return_final}
	\bar{G}^{m,l,y}_k=\tilde{G}^{m,l,y}_k + \sum^{k+d^{l,m}_{prev}+\Delta d-1}_{i=k+d^{l,m}_{prev}}r(b_i,\pi_{rollout}(b_i), b_{i+1}).
\end{align}
$d^{l,m}_{prev}$ is the horizon of propagated belief $b^{-l,m}_{k_{m}+1}$,
$\bar{G}^{m,l,y}_k$  shares the same values in the summation as the return $\tilde{G}^{m,l,y}_k$ \eqref{eq:correct_g_reuse},
but it also includes additional terms from the extended trajectory due to the horizion extension using the rollout policy $\pi_{rollout}$. Therefore, only the rewards for these additional terms need to be computed when extending the horizon, while all shared terms ($\tilde{G}^{m,l,y}_k$) can be reused.

Since the tree policy varies between simulations, the update represented by \eqref{eq:mis_est_prop_belief_alg} operates in a heuristic manner, with its convergence yet to be established. We intend to explore this aspect in a future work. 

After extending the horizon of a reused propagated belief node $b^-$ and reusing its action-value function, the counter $N(b_k,a_k)$ is incremented by the visitation count $N(b^-)$ using the relation $N(b_k,a_k)=\sum_{b^-_k \in C(b_k,a_k)}N(b^-_k)$.

This approach accelerates our algorithm for a given number of simulations, offering a speedup over PFT-DPW \cite{Sunberg18icaps}.

To summarize, here is the high-level logical flow of our algorithm:
At each iteration, we either reuse a propagated belief node by extending its horizon by $\Delta d$, as illustrated in Figure \ref{fig:horizon_gap_illustration}, and compute the extended return for the subtree rooted at the reused node $b^-$, or create a new node. 
Subsequently, the Multiple Importance Sampling (MIS) estimator \eqref{eq:mis_est_prop_belief_alg} is employed to evaluate the action-value function. To avoid the computational expense of a naive calculation, we leverage Theorem \ref{th:incremental_mis_update} to perform efficient incremental updates of the estimator in \eqref{eq:mis_est_prop_belief_alg}.

\subsection{Algorithm Description}
The complete algorithm is outlined across multiple methods - Algs.~\ref{alg:general_planning_loop_new}, \ref{alg:main_alg_new}, \ref{alg:reuse_new} and \ref{alg:add_horizon_layers_new}. 
Alg.~\ref{alg:general_planning_loop_new} illustrates a general planning loop wherein the agent iteratively plans and executes actions until the problem is solved. 
After each planning session, reuse candidates are updated based on the preceding planning tree.
\begin{algorithm}
	\caption{General Planning Loop}
	\label{alg:general_planning_loop_new}
	\begin{algorithmic}[1] 
		\STATE \textbf{Procedure}: \textsc{Solve}(b, D)
		\WHILE{ProblemNotSolved()}
		\STATE $a  \leftarrow Plan(b,D)$ 
		\STATE $o \leftarrow ReceiveObservation(b,a)$
		\STATE $b',b'^-,r \leftarrow G_{PF(m)}(b,a,o)$
		\STATE UpdateReuseCandidates($a, D, b, b')$
		\STATE $b \leftarrow b' $
		\ENDWHILE
	\end{algorithmic}
\end{algorithm}
The main algorithm is detailed in Alg.~\ref{alg:main_alg_new} with key modifications compared to the PFT-DPW algorithm \cite{Sunberg18icaps}
highlighted in \textcolor{red}{red}. 
The ActionProgWiden method (line \ref{line:action_pw}) is implemented following the same approach as described in \cite{Sunberg18icaps}.  
ShouldReuse method (line \ref{line:should_reuse}) evaluates three conditions: current node $b$ is the root, the balance between reused and new nodes, and the availability of reuse candidates. 
The second criterion is important because, while acquiring estimates from prior partial trajectories is runtime-efficient, generating new trajectory samples from the correct distribution is essential.
Currently, our algorithm only applies reuse to the root node, as it promises the most significant computational savings. Since the root node typically has the shallowest depth in the tree, we can optimize by conserving numerous reward computations for most of its descendants.
While extending reuse to nodes at other depth levels is feasible, it falls outside the scope of this work.
 
The GetReuseCandidate method (line \ref{line:get_reuse_candidate}) selects a reuse candidate propagated belief $b'^-$ from the dataset $D$ based on a distance function $f_D$ (line \ref{line:get_closest_prop_belief}). An example of $f_D$ is $||\Exp[b^- - b^-_{MLE}]||^2_2$ where $b^-_{MLE}$ represents the maximum likelihood propagated belief, given belief $b$ and action $a$ 
which can be calculated using \eqref{propagated_belief_likelihood} 
with $O(m)$ complexity where $m$ is the number of samples.
Since $f_D$ is applied to the entire dataset, it needs to be computationally efficient. Additionally, reusing nodes with high visitation counts will further reduce the overall runtime of the algorithm. 

\begin{algorithm}
	\caption{IR-PFT}
	\label{alg:main_alg_new}
	\begin{algorithmic}[1] 
		\STATE \textbf{Procedure}: \textsc{Plan}(b, D)
		\STATE $i = 0$
		\WHILE{$i < n$}
		\STATE $Simulate(b, d_{max}, D)$
		\ENDWHILE
		\STATE $a$ = $argmax_a\{Q(b,a)\}$
		\STATE return $a$
		\item[]
	\end{algorithmic}
	
	\begin{algorithmic}[1]
		\STATE \textbf{Procedure}: \textsc{Simulate}(b, d, D)
		\IF {$d = 0$}
		\STATE return 0
		\ENDIF
		\STATE $a \leftarrow $ ActionProgWiden($b$)\label{line:action_pw}
		\IF{$|C(ba)|\leq k_oN(ba)^{\alpha_o}$}
		\IF{\textcolor{red}{$ShouldReuse(b,a, D)$}} \label{line:should_reuse}
		\STATE \textcolor{red}{$\br \leftarrow  GetReuseCandidate(b,a,D)$} \label{line:get_reuse_candidate}
		\STATE \textcolor{red}{FillHorizonPropagated($\br,d-d_{b^-}$)} \label{line:horizon_alignment}
		\STATE \textcolor{red}{$N(b) \leftarrow N(b) + N(\br)$} \label{line:belief_counter}
		\STATE \textcolor{red}{$N(ba) \leftarrow N(ba) + N(\br)$} \label{line:belief_action_counter}
		\STATE \textcolor{red}{$i \leftarrow i +  N(\br)$} \label{line:update_simulation_counter}
		\COMMENT{update simulation counter}
		\STATE \textcolor{red}{$Q(ba)\leftarrow MISUpdate()$} \label{line:mis_update_first} \COMMENT{update using  \eqref{eq:mis_est_prop_belief_alg}}
		\STATE \textcolor{red}{$C(b,a) \leftarrow C(b,a) \cup \{(\br)\}$ } \label{line:save_prop_belief_first}
		\STATE \textcolor{red}{return $total$}
		\ELSE
		\STATE \textcolor{red}{$b',b'^-, r \leftarrow G_{PF(m)}(b,a)$} \label{line:pf_update}
		\STATE \textcolor{red}{$C(b,a) \leftarrow C(b,a) \cup \{(b'^-)\}$ }
		\STATE \textcolor{red}{$C(b'^-) \leftarrow C(b'^-) \cup \{(b',r)\}$} \label{line:save_prop_belief_second}
		\STATE $total \leftarrow r + \gamma ROLLOUT(b',d-1)$
		\ENDIF
		\ELSE
		\STATE \textcolor{red}{$b'^- \leftarrow$ sample uniformly from $C(ba)$} \label{line:uniform_sample_prop}
		\STATE \textcolor{red}{$b',r \leftarrow$ sample uniformly from $C(b'^-)$} \label{line:uniform_sample_posterior}
		\STATE $total \leftarrow r + \gamma Simulate(b',d-1,T)$
		\ENDIF
		\STATE $N(b) \leftarrow N(b)+1$
		\STATE $N(ba) \leftarrow N(ba)+1$
		\STATE \textcolor{red}{$Q(ba)\leftarrow MISUpdate()$} \label{line:mis_update_second} \COMMENT{update using  \eqref{eq:mis_est_prop_belief_alg}}
		\STATE return $total$
	\end{algorithmic}
\end{algorithm}

\begin{algorithm}[tb]
	\caption{Reuse Functions}
	\label{alg:reuse_new}
	\begin{algorithmic}[1] 
		\STATE \textbf{Procedure}: \textsc{UpdateReuseCandidates}(a,D,$b_k$,$b^{real}_{k+1}$)
		\STATE ReuseDict $dict \leftarrow$ $\{\}$
		\FOR{$b^-_{k+1} \in C(b_{k},a)$}
		\FOR{$b_{k+1}\in C(b^-_{k+1})$}
		\FOR{$a' \in Actions(b_{k+1})$}
		\FOR{$b^-_{k+2} \in C(b_{k+1},a')$}
		\IF{$n(b^-_{k+2}) > n_{min}$}
		\STATE $D.append(b^-_{k+2})$
		\ENDIF	
		\ENDFOR	
		\ENDFOR	
		\ENDFOR	
		\ENDFOR
		\item[]
	\end{algorithmic}
	\begin{algorithmic}[1]
		\STATE \textbf{Procedure}: \textsc{ShouldReuse}(b,a,D)
		\IF{not b.IsRoot()}
		\STATE return $false$
		\ENDIF
		\IF{$NumReused(b,a) > \frac{Total(b,a)}{2}$}
		\STATE return $false$
		\ENDIF
		\STATE $candidates \leftarrow D.GetReuseCandidatesDict()$
		\STATE return not($candidates.empty()$)
		\item[]
	\end{algorithmic}
	
	\begin{algorithmic}[1]
		\STATE \textbf{Procedure}: \textsc{GetReuseCandidate}(b,a,D)
		\STATE $b^- \leftarrow argmin_{b^-}\{f_D(b^-,b,a)\}$ \label{line:get_closest_prop_belief}
		\STATE return $b^-$ 
	\end{algorithmic}
\end{algorithm}

The FillHorizonPropagated method (line \ref{line:horizon_alignment}), addresses discrepancies in horizon lengths when reusing nodes from the previous planning sessions. Algorithm \ref{alg:add_horizon_layers_new} performs recursive traversal of the subtree defined by propagated belief $b'^-$ and extends its depth by d using the rollout policy.

At lines \ref{line:belief_counter} and \ref{line:belief_action_counter}, we increment counters, where $N(\br)$ represents the count of trajectories passing through reuse candidate propagated belief node $\br$.
At line \ref{line:update_simulation_counter}, we increment the \textsc{Plan} procedure counter by $N(\br)$.

At lines \ref{line:mis_update_first} and \ref{line:mis_update_second} we utilize \eqref{eq:mis_est_prop_belief_alg} to update $Q(b,a)$, leveraging efficiency through the application of Theorem \eqref{th:incremental_mis_update}. 
At line \ref{line:save_prop_belief_first} we store the propagated belief $\br$.

Lines \ref{line:pf_update} - \ref{line:save_prop_belief_second} are executed when we choose not to reuse and instead initialize a new propagated belief from scratch. A new belief is generated using the particle filter method \cite{Thrun05book}, after which the propagated belief and posterior belief are saved, and a rollout is performed.

At lines \ref{line:uniform_sample_prop} and \ref{line:uniform_sample_posterior} we sample uniformly both propagated and posterios beliefs.

UpdateReuseCandidates method in Algorithm \ref{alg:reuse_new} inserts new reuse candidates that have a visitation count larger than a threshold $n_{min}$, as we aim to reuse nodes with higher visitation counts, which leads to a greater speedup.

\begin{algorithm}
	\caption{Fill Horizon Gap}
	\label{alg:add_horizon_layers_new}
	\begin{algorithmic}[1] 
		\STATE \textbf{Procedure}: \textsc{FillHorizonPropagated}($b^-,d$)
		\STATE $Q_{new}(b^-) \leftarrow 0$
		\FOR{$b' \in C(b^-)$}
		\STATE $Q_{new}(b^-) \leftarrow Q_{new}(b^-) + FillHorizonPosterior(b')$
		\ENDFOR	
		\STATE return $\frac{Q_{new}(b^-)}{|C(b^-)|}$
		\item[]
	\end{algorithmic}
	
	\begin{algorithmic}[1] 
		\STATE \textbf{Procedure}: \textsc{FillHorizonPosterior}($b,d$)
		\IF{$IsLeaf(b)$}
		\STATE $a \leftarrow DefaultPolicy(b)$
		\STATE $b',b'^-,r \leftarrow G_{PF(m)}(b,a)$
		\STATE $N(b,a) \leftarrow 1$
		\STATE $N(b'^-) \leftarrow 1$
		\STATE $Q(b'^-) \leftarrow r$
		\STATE $Q(b,a) \leftarrow r$
		\STATE return $r$
		\ENDIF
		\STATE $Q(b) \leftarrow 0$
		\FOR{$a \in Actions(b)$}
		\FOR{$b'^- \in C(b,a)$}
		\STATE $Q(b) \leftarrow Q(b) + FillHorizonPropagated(b'^-, d-1)$
		\ENDFOR	
		\ENDFOR	
		\STATE $Q(b) \leftarrow \frac{Q(b)}{N(b)}$
		\STATE return $Q(b)$
		\item[]
	\end{algorithmic}
\end{algorithm}

\section{RESULTS}
\begin{figure*}[htbp] 
	\centering
	\begin{subfigure}[b]{0.2\textwidth}
		\includegraphics[width=\textwidth]{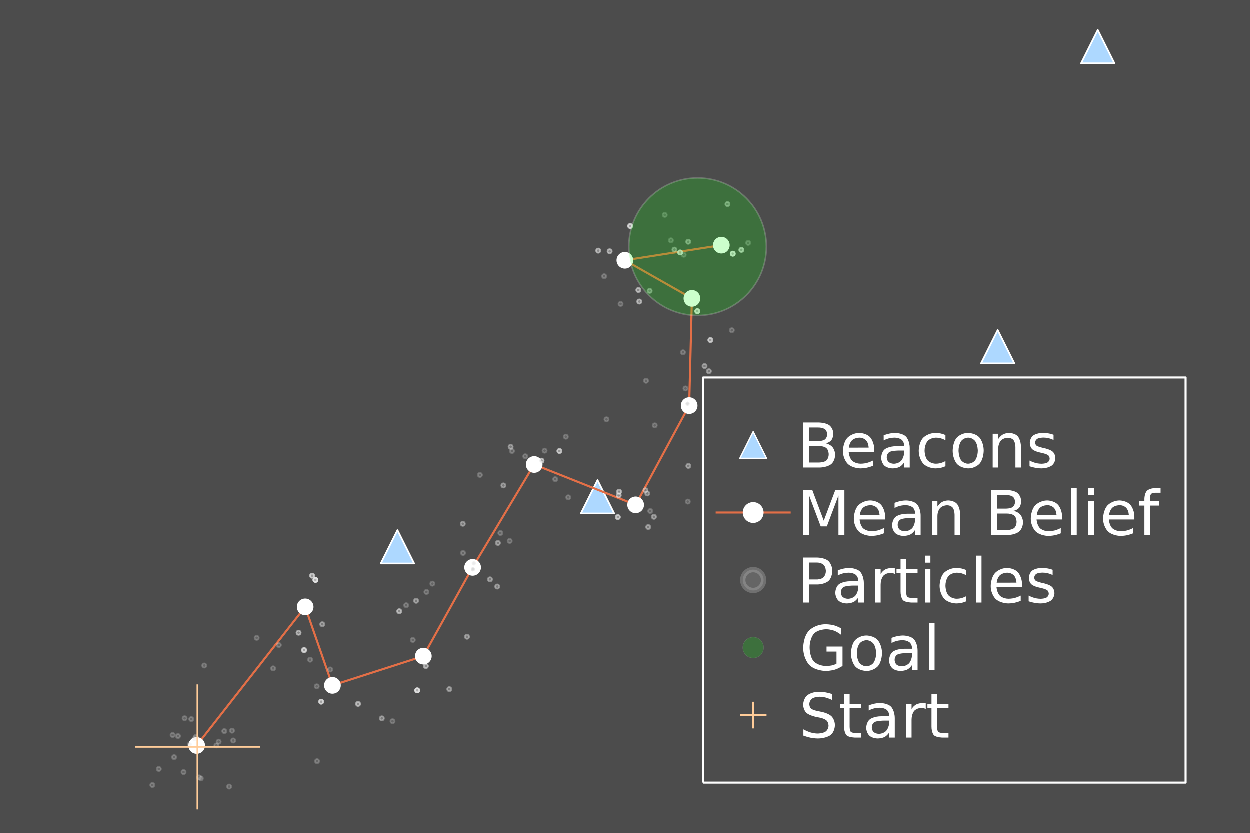}
		\caption{Illustration of the continuous light dark 2d environment}
		\label{fig:light_dark}
	\end{subfigure}
	\begin{subfigure}[b]{0.23\textwidth}
		\includegraphics[width=\textwidth]{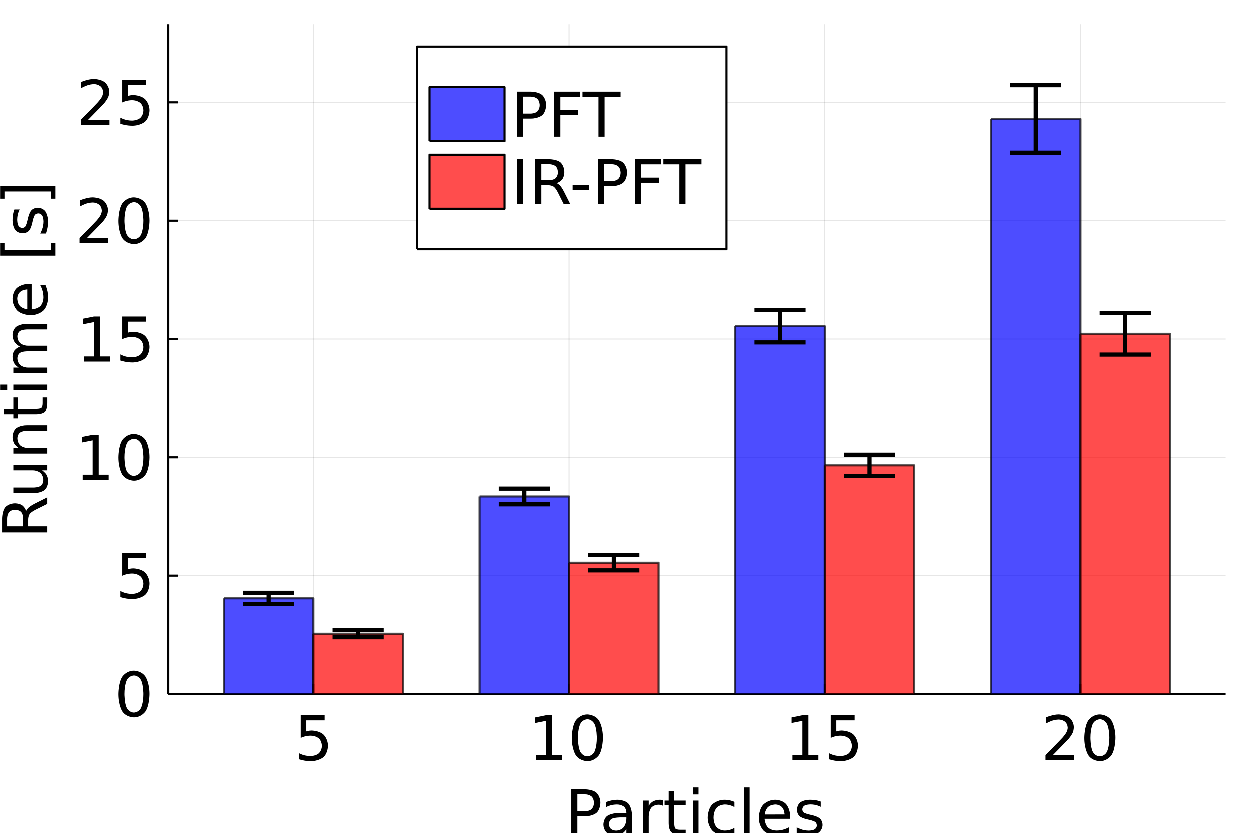}
		\caption{Runtime comparison}
		\label{fig:runtime_improvement}
	\end{subfigure}
	\hfill 
	\begin{subfigure}[b]{0.23\textwidth}
		\includegraphics[width=\textwidth]{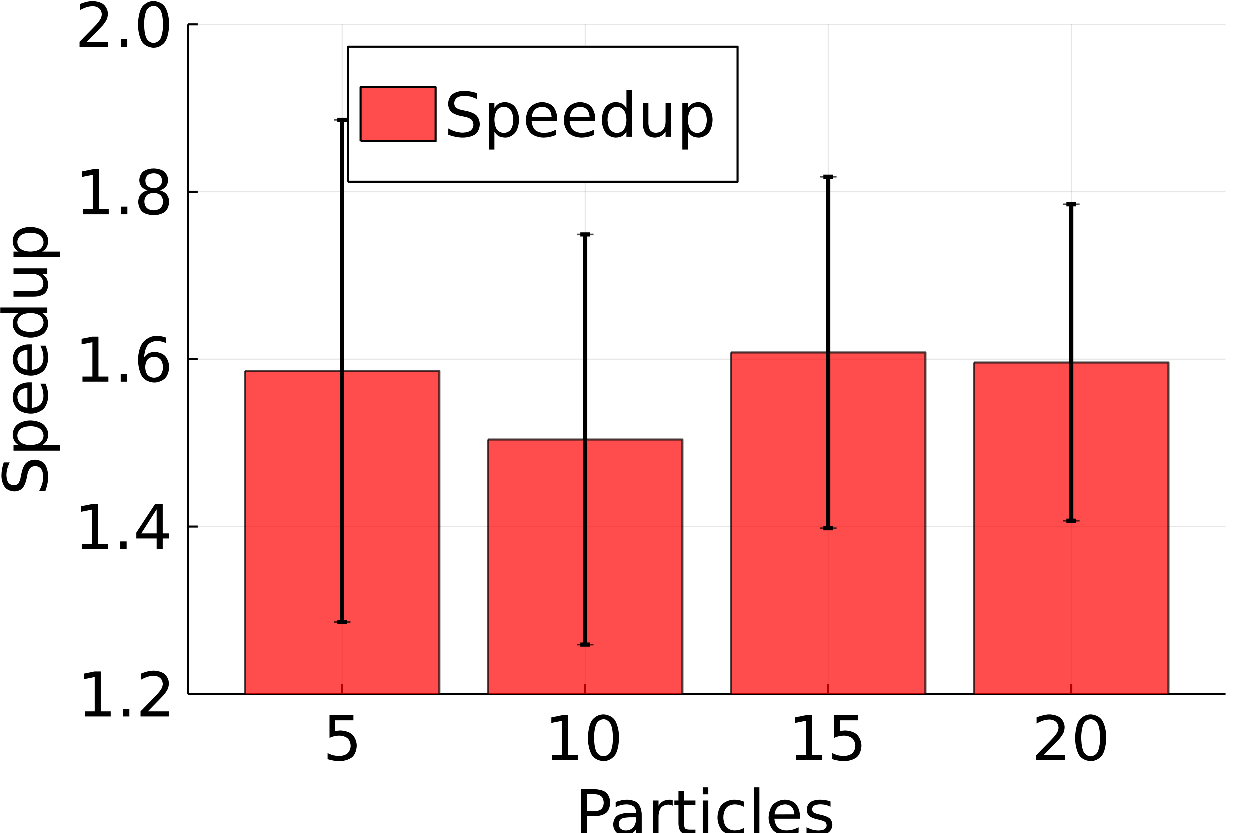}
		\caption{Speedup}
		\label{fig:speedup_improvement}
	\end{subfigure}
	\hfill 
	\begin{subfigure}[b]{0.23\textwidth}
		\includegraphics[width=\textwidth]{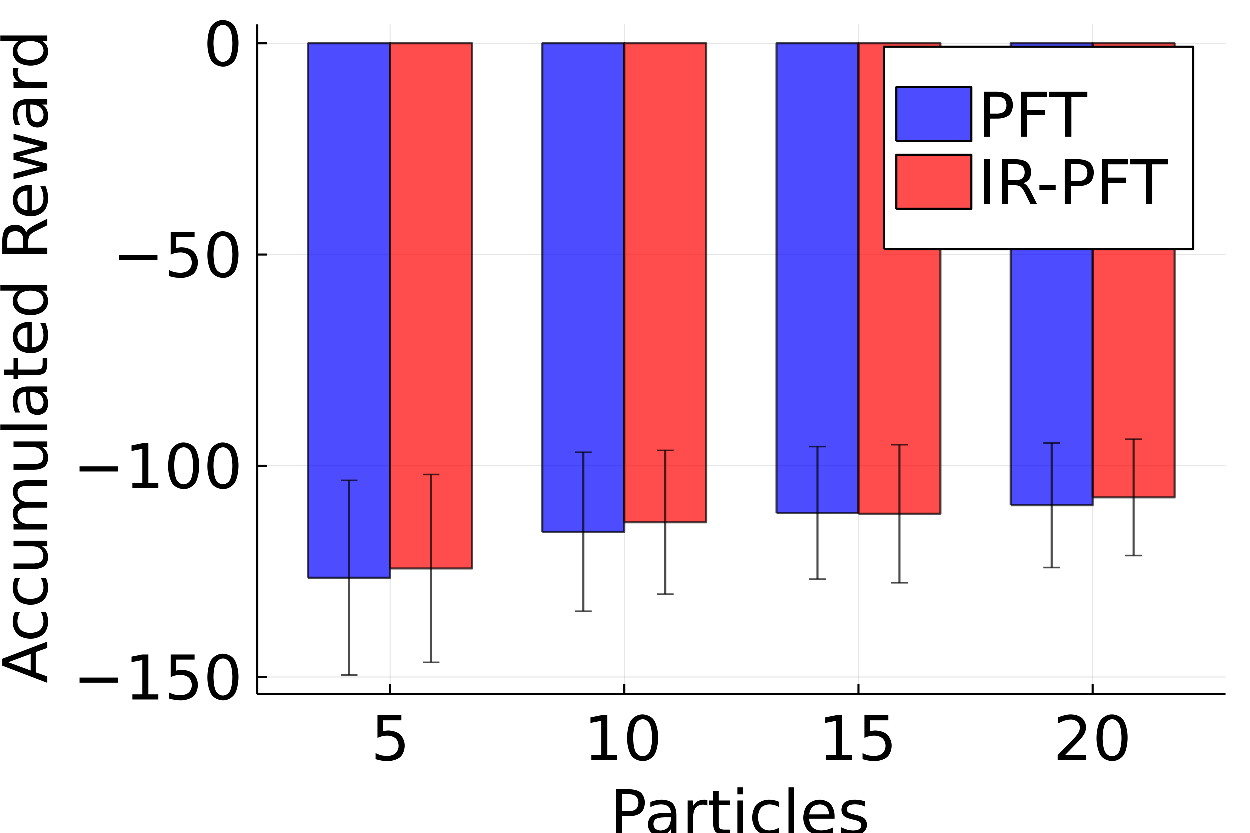}
		\caption{Accumulated reward comparison}
		\label{fig:accumulated_reward}
	\end{subfigure}
	\caption{Light dark experiments comparing PFT and IR-PFT.}
	\label{fig:threeplots}
\end{figure*}

We assess the performance of the IR-PFT algorithm by comparing it to the PFT-DPW algorithm \cite{Sunberg18icaps}. Our evaluation focuses on two main aspects: runtime and accumulated reward, with statistics measured for each.
In all experiments, the solvers were limited to 1000 iterations for each planning phase. All experiments were conducted using the standard 2D Light Dark benchmark, wherein the agent's objective is to reach the goal while simultaneously minimizing localization uncertainty by utilizing beacons distributed across the map; refer to the illustration in Figure \ref{fig:light_dark}. The reward function is defined as a weighted sum of the average distance to goal and differential entropy estimator which is calculated using \cite{Boers10fusion}. Each algorithm was evaluated using different quantities of particles—specifically, $5, 10, 15$, and $20$, while maintaining a constant horizon length of $d=10$.  In the following results reuse was done according to ShouldReuse method (line \ref{line:should_reuse}) as detailed in the previous section.
We compared the runtime of IR-PFT with and without reuse vs PFT-DPW. The results are depicted in Figure \ref{fig:runtime_improvement} as a function of number of particles.
Additionally, we included a speedup chart, which provides more insightful information, in Figure \ref{fig:speedup_improvement}. 
The runtime of IR-PFT consistently outperformed that of PFT. We observed a saturation in the speedup at a factor of approximately 1.5, which is attributed to the savings in reward computation, the most computationally intensive part of algorithm.
We compare the accumulated rewards of IR-PFT with and without reuse (Figure \ref{fig:accumulated_reward}). The results show negligible differences, indicating that our method improves runtime without compromising planning performance.

\section{Conclusions and Future Work}
In this paper, we have proposed a general framework which allows to reuse prior information during the current planning session. 
We derived theoretical justification for reuse via Multiple Importance Sampling and introduced a new MCTS-like algorithm, IR-PFT which reuses information from previous planning session and allows to speed up calculations in current planning session.
In order to evaluate IR-PFT algorithm, we conducted an empirical performance study. Specifically, we compared the performance of our approach with and without the reuse of prior information.
We measured various performance metrics, including computation time and the accumulated reward. Our results clearly indicate a speed-up in the planning process when prior information is leveraged. Importantly, despite the accelerated computations, our approach maintains the same level of performance as the traditional planning approach without reuse.
Incorporating prior information significantly boosts planning efficiency, delivering time savings while maintaining high-quality results. These findings underscore the effectiveness and potential of the proposed approach.

\appendix 
\subsection{Proof of Theorem \ref{th:incremental_mis_update}}\label{th1_proof}
Consider an MIS estimator \eqref{mis_balance_h} with $M$ different distributions and $n_m$ samples for each distribution $q_m \in \{q_1,..., q_M\}$. Given a  batch of $L$ I.I.D. samples from distribution $q_{m'}$ which may be an existing or new distribution, $$\hat{\Exp}^{MIS}_p[f(x)]=\sum^{M}_{m=1}\sum^{n_m}_{i=1}\frac{p(x_{i,m})}{\sum^M_{j=1} n_j \cdot q_{j}(x_{i,m})} f(x_{i,m}),$$ \eqref{mis_balance_h} can be efficiently updated with a computational complexity of O($M\cdot n_{avg} + M \cdot L$) and memory complexity O($M \cdot n_{avg}$).

For every distribution $q_m$ we have the term $\sum^{n_m}_{i=1}\frac{p(x_{i,m})}{\sum^M_{j=1} n_j \cdot q_{j}(x_{i,m})}f(x_{i,m})$. 

In case $m \neq m'$:

$\sum^M_{j=1} n_j \cdot q_{j}(x_{i,m}) \leftarrow \sum^M_{j=1}  n_j \cdot q_{j}(x_{i,m}) + L\cdot q_{m'}(x_{i,m})$ - $O(1)$ complexity.
We have $n_m$ samples and $M$ distributions so the complexity of this update is $O(M\cdot n_m)$.

In case $m = m'$:

For existing samples
$\sum^M_{j=1}  n_j \cdot q_{j}(x_{i,m}) \leftarrow \sum^M_{j=1}  n_j \cdot q_{j}(x_{i,m}) + L\cdot q_{m'}(x_{i,m})$ - $O(1)$ complexity.
We have $n_m$ samples existing samples so in total $O(n_m)$ complexity.

For each new sample we need to calculate $\frac{p(x_{i,m})}{\sum^M_{j=1}  n_j \cdot q_{j}(x_{i,m})} f(x_{i,m})$ - $O(M)$ complexity.
We have $L$ new samples so in total $O(L \cdot M)$ complexity.
The total complexity of the update is $O(M \cdot n_{avg} + M \cdot L)$.
\subsection{Proof of Theorem \ref{th:trajectory_likelihood_rat}}\label{th:trajectory_likelihood_rat_proof}
\begin{align} \label{eq_proof:traj_likelihood_ratio}
	\frac{\prob(\TRE{i}|b_k, a_k, \pi)}{\prob(\TRE{i}|b^{i}_{k_i},a^i_{k_i}, \pi)} =
	\frac{\prob(b^{-i}_{k_i+1}, ... ,b^{i}_{k_i+L}|b_k, a_k,\pi)}{\prob(b^{-i}_{k_i+1}, ... , b^i_{k_i+L}|b^{i }_{k_i}, a^i_{k_i}, \pi)}.
\end{align}
Applying chain rule yields,
\begin{align}
	\frac{\prob(b^{-i}_{k_i+1}|b_k, a_k)}{\prob(b^{-i}_{k_i+1}|b^{i}_{k_i}, a^i_{k_i})} & \cdot \cancel{\frac{\prob(o^i_{k_i+1}, \dots, b^{i}_{k_i+L}|b^{-i}_{k_i+1}, \pi)}{\prob(o^i_{k_i+1}, \dots, b^{i}_{k_i+L}|b^{-i}_{k_i+1}, \pi)}} = \\
	&= \frac{\prob(b^{-i}_{k_i+1}|b_k, a_k)}{\prob(b^{-i}_{k_i+1}|b^{i}_{k_i}, a^i_{k_i})} \notag.
\end{align}

\addtolength{\textheight}{-12cm}   



\end{document}